\documentclass[10pt,conference]{IEEEtran}
\IEEEoverridecommandlockouts
\usepackage{cite}
\usepackage{amsmath,amssymb,amsfonts}
\usepackage{algorithmic}
\usepackage{graphicx}
\usepackage{textcomp}
\usepackage{xcolor}
\usepackage{amsmath,lipsum}
\usepackage[ruled]{algorithm2e}
\usepackage{graphicx}
\usepackage{float}
\usepackage{subfigure}
\usepackage{multirow}
\usepackage{multicol}
\usepackage{arydshln}
\usepackage{cite}
\usepackage{stfloats}
\usepackage{booktabs}
\usepackage{caption}

\IEEEoverridecommandlockouts\IEEEpubid{\makebox[\columnwidth]{ 978-1-6654-3540-6/22~\copyright~2022 IEEE \hfill} \hspace{\columnsep}\makebox[\columnwidth]{ }}

\def\BibTeX{{\rm B\kern-.05em{\sc i\kern-.025em b}\kern-.08em
    T\kern-.1667em\lower.7ex\hbox{E}\kern-.125emX}}
\begin{document}

\title{Graph Reinforcement Learning-based CNN Inference Offloading in Dynamic Edge Computing\\ }
\author{\IEEEauthorblockN{Nan~Li, Alexandros~Iosifidis and Qi~Zhang }
\IEEEauthorblockA{DIGIT, Department of Electrical and Computer Engineering, Aarhus University.\\
Email: \{linan, ai, qz\}@ece.au.dk}}
\maketitle

\begin{abstract}
This paper studies the computational offloading of CNN inference in dynamic multi-access edge computing (MEC) networks. To address the uncertainties in communication time and Edge servers' available capacity, we use early-exit mechanism to terminate the computation earlier to meet the deadline of inference tasks. We design a reward function to trade off the communication, computation and inference accuracy, and formulate the offloading problem of CNN inference as a maximization problem with the goal of maximizing the average inference accuracy and throughput in long term. To solve the maximization problem, we propose a graph reinforcement learning-based
early-exit mechanism (GRLE), which outperforms the state-of-the-art work, deep reinforcement learning-based online offloading (DROO) and its enhanced method, DROO with early-exit mechanism (DROOE), under different dynamic scenarios. The experimental results show that GRLE achieves the average accuracy up to 3.41$\times$ over graph reinforcement learning (GRL) and 1.45$\times$ over DROOE, which shows the advantages of GRLE for offloading decision-making in dynamic MEC. 
\end{abstract}

\begin{IEEEkeywords}
Dynamic computation offloading, CNN inference, Graph reinforcement learning, Edge computing, Service reliability
\end{IEEEkeywords}

\section{Introduction}
The advancement in convolutional neural networks (CNNs) has propelled various emerging CNN-based IoT applications, such as autonomous driving and augmented reality. For time-critical IoT applications, not only high reliability but also low latency is crucial to enable reliable services and safe intelligent control \cite{008}. To achieve higher inference accuracy, deeper CNNs with massive multiply-accumulate operations are often required. However, resource-limited IoT devices are not feasible to complete computational intensive CNNs within a stringent deadline \cite{Nan2022ICC}. 

Edge computing is a paradigm that allows IoT devices to offload computational tasks to their nearby edge servers (ESs) via wireless channels \cite{008, Nan2022}. However, the stochastic wireless channel states may cause fluctuations in communication time, which consequently results in that the exact task completion time is unknown in advance and varies over time \cite{Nan2017}. To address the uncertainties in communication time, dynamic offloading methods are proposed to strike a balance between the communication and computation \cite{Guo2016,Tran2019,Li2018,Huang2020}. However, when communication takes too much time or the available computation resource at ESs is insufficient, it cannot meet the stringent deadline if running an inference task through the entire pre-trained CNN model (e.g., until the end of the main branch of a CNN). To address this issue, dynamic inference
methods \cite{bakhtiarnia2022dynamic} are promising to meet the latency requirements through modification of CNN architecture and allowing dynamic inference time at the compromise of inference accuracy.
\begin{figure}
    \centering
    \includegraphics[width=0.35\textwidth]{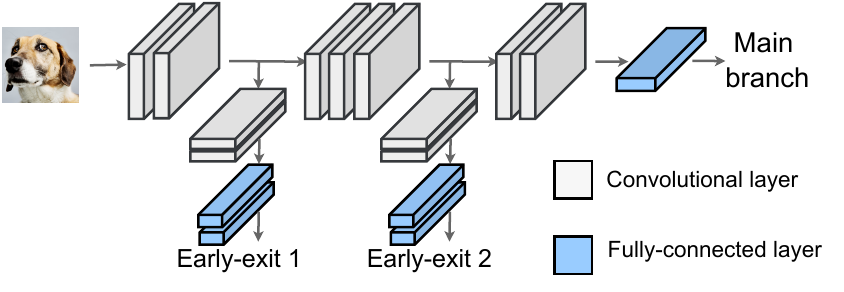}
    \vspace{-1mm}
    \caption{Early-exit mechanism of CNN inference}
    \label{fig:early-eixt}
    \vspace{-5mm}
\end{figure}

Early-exit mechanism \cite{Arian2021} is a dynamic inference method, which terminates the inference at an early stage, as shown in Fig. \ref{fig:early-eixt}. Unlike the conventional CNN inference, early-exit architecture enables the intermediate classifiers of a CNN to provide different accuracy-latency trade-offs along its depth. The goal of early-exit architecture is to provide adaptive accuracy-latency behavior so that each inference task terminates at an appropriate exit based on its computation time budget. In other words, in case of poor channel states, an early-exit can be used to terminate the computation earlier, thereby providing the inference result within the deadline of time critical applications at the expense of slightly lower accuracy, instead of completely missing the deadline. However, how to design an efficient task offloading scheme to trade off communication, computation and inference accuracy in dynamic multi-access edge computing (MEC) is a challenging problem and has not yet been adequately addressed.  

Based on the motivation above, this paper studies the offloading problem of CNN inference tasks in dynamic MEC networks. Our contributions are summarized as follows.
\begin{itemize}
	\item We use early-exit CNN architecture to provide dynamic inference to address the issue caused by stochastic available computing resource of ESs and the uncertainties in communication time, thereby ensuring that inference task is completed within time constraints. 
	\item We define a reward function to strike a balance among the communication, computation and inference accuracy. Then we model the CNN inference offloading problem as a maximization problem to maximize the average inference accuracy and throughput in long term.
	\item We design a graph reinforcement learning-based early-exit mechanism (GRLE) to make optimal offloading decisions. The experimental results show that GRLE achieves better performance in comparison with the state-of-art work, deep reinforcement learning-based online offloading (DROO) \cite{Huang2020} and its enhanced method, DROO with early-exit (DROOE), under different dynamic scenarios. In our experiments, GRLE achieves average accuracy up to 3.41$\times$ over graph reinforcement learning (GRL) and 1.45$\times$ over DROOE, which demonstrates that GRLE is effective to make offloading decision in dynamic MEC.
\end{itemize}
\section{Related work}\label{section:system_model}
Due to the characteristics of time-varying wireless channels, it is crucial to make effective offload decisions to ensure the QoS in edge computing paradigms. Guo et al. \cite {Guo2016} proposed a heuristic search-based energy-efficient dynamic offloading scheme to minimize energy consumption and completion time of task with strict deadline. Tran et al. \cite{Tran2019} designed a heuristic search method to optimize task offloading and resource allocation to jointly minimize task completion time and energy consumption by iteratively adjusting the offloading decision. However, heuristic search methods require accurate input information, which is not applicable to dynamic MEC.

Reinforcement learning (RL) is a holistic learning paradigm that interacts with the dynamic MEC to maximize long-term rewards. Li et al. \cite{Li2018} proposed to use deep RL (DRL)-based optimization methods to address dynamic computational offloading problem. However, applying DRL directly to the problem is inefficient in a practical deployment because offloading algorithms typically require many iterations to search an effective strategy for unseen scenarios. Huang et al. \cite{Huang2020} proposed DROO to significantly improve the convergence speed through efficient scaling strategies and direct learning of offloading decisions. However, the DNN used in DROO can only handle Euclidean data, which makes it not well suitable for the graph-like structure data of MEC. In addition, all the above methods do not provide dynamic inference, which is lack of flexibility in making good use of any available computation resource under stringent latency. 
\section{system model}\label{section:system_model}
This paper studies the computation offloading in an MEC network, which consists of $M$ IoT devices and $N$ ESs. The set of IoT devices and ESs are denoted as $\mathbb{M}
= \left\{ 1, 2, \cdots, M \right\}$ and $\mathbb{N} = \left\{ 1, 2, \cdots, N \right\}$ respectively. At each time slot $k \in \mathbb{K}$, each IoT device generates a computational task that has to be processed within its deadline, where $\mathbb{K} = \left\{ 1, 2, \cdots, K \right\}$. The length of each time slot $k$ is assumed to be $\tau$. The computational task of IoT device is assumed to use a CNN with $L$ convolutional layers (CLs), which are denoted as $\mathbb{L} = \left\{ 1, 2, \cdots, L \right\}$. Each IoT device can offload its inference task to one ES through wireless channel and each ES can serve multiple IoT devices at each time slot. In the case of poor wireless channel state or insufficient available computation resource of ESs, ESs can use the early-exit mechanism to terminate the computation earlier to meet the deadline of an inference task. After completing inference task, ES will send back the results to IoT devices. To perform computational offloading, two decisions should be considered: (1) to which ES an IoT device should offload its tasks; (2) to which early-exit the ES can perform the task based on a time constraint.
\subsection{Communication time}
Computation offloading involves delivering of task data and its inference result between IoT device and ES. We assume that a task of IoT device $m$ is offloaded to ES $n$ at time slot $k$. The task information is expressed as $\theta_{m,n}^k = \left\{ d_{m,n}^k,\delta_{m,n}^k, r_{m,n}^k \right\}$, where $d_{m,n}^k$ and $\delta_{m,n}^k$ are the task size and latency requirement respectively, and $r_{m,n}^k$ is the uplink transmission data rate from IoT device $m$ to ES $n$. Therefore, the transmission delay of an inference task from IoT device $m$ to ES $n$ is denoted as
\setcounter{equation}{0}
\begin{equation}
t_{m,n}^{\textit{com}}\left(k \right) = \alpha_{m,n}^k d_{m,n}^k / r_{m,n}^k
\label{eq2}
\end{equation}
where $\alpha_{m,n}^k \in \left\{0,1\right\}$ is a binary variable to indicate whether the task of device $m$ is offloaded to ES $n$ at time slot $k$. As each inference task can only be offloaded to one ES, there is
\begin{equation}
    \sum\limits_{n \in \mathbb{N}} \alpha_{m,n}^k = 1, \forall m \in \mathbb{M}.
    \label{eq2}
\end{equation}

In addition, as the output of CNN is very small, often a number or value that
represents the classification or detection result, transmission delay of the feedback is negligible.
\subsection{Computation time}
During computation, an ES can only select an early-exit to perform inference for each offloaded task. As we use a binary variable
$\beta_{m,n,l}^k \in \left\{0,1\right\}$ to denote if ES $n$ performs device $m$'s task until early-exit $l$, there is
\begin{equation}
    \sum\limits_{l \in \mathbb{L}} \beta_{m,n,l}^k = 1, \forall m \in \mathbb{M} \; {\rm{and}} \; n \in \mathbb{N}.
        \label{eq3}
\end{equation}

Assuming that ES $n$ performs an inference task until early-exit $l$, the computation time and inference accuracy are denoted as $t_{n,l}^{\textit{cmp}}$ and  $\phi_{l}$ respectively. Therefore, the computation time of $m$'s task on ES $n$ is expressed as 
\begin{equation}
    t_{m,n}^{\textit{
    cmp}}(k) = \beta_{m,n,l}^k t_{n,l}^{\textit{cmp}}.
\label{eq4}
\end{equation}
Correspondingly, the achieved inference accuracy of the task generated by IoT device $m$ at time slot $k$ is 
\begin{equation}
    \Phi_{m,n}(k) = \beta_{m,n,l}^k \phi_{l}.
\label{eq4}
\end{equation}

We assume each ES processes inference tasks on a first-come-first-served basis. Namely, ES $n$ can start to process a new arrival task only if it has processed all the previously arrived tasks. Assume that device $m$'s task generated at time slot $k$ is offloaded to ES $n$, device $m$ can start transmission of this task only after completing transmission of its previous tasks. Since the propagation time is negligible, the task generated by device $m$ at time slot
$k$ arrives at ES $n$ at time instant, $T_{m,n}^{\textit{a}}\left(k\right)$, which can be expressed as
\begin{equation}
\small
T_{m,n}^{\textit{a}}\hspace{-1mm}\left(k\right) \hspace{-1mm}=\hspace{-1mm} \left\{ {\begin{array}{*{20}{c}}
\hspace{-2mm}t_{m,n}^{\textit{com}}\left(k\right) &{k\hspace{-1mm} = \hspace{-1mm}1},\\
    \hspace{-2mm}\max \hspace{-0.5mm} \left(T_{m,n'}^{\textit{a}}\hspace{-1mm}\left(k\hspace{-1mm}-\hspace{-1mm}1 \hspace{-0.5mm}\right)\hspace{-0.5mm},\hspace{-0.5mm} \left(k\hspace{-1mm}-\hspace{-1mm}1\right)\hspace{-0.5mm}\tau\hspace{-0.5mm}\right) \hspace{-0.5mm}+\hspace{-0.5mm}  t_{m,n}^{\textit{com}}\hspace{-0.5mm}\left(k\right)&{k \hspace{-1mm}\neq \hspace{-1mm}1, n'\hspace{-1mm}\in\hspace{-1mm}\mathbb{N}}. 
\end{array}} \right.
\label{eq5}
\end{equation}
Correspondingly, the waiting time at ES $n$ of device $m$'s task generated at time slot $k$, $t_{m,n}^{\textit{w}}\left(k\right)$, can be calculated as (7), where ${\bf{1}}\left(\cdot \right)$ is an indicator function to show the occurrence of device $m'$'s task arriving at ES $n$ before device $m$'s task.
\begin{figure*}[t]
    \centering
    \includegraphics[width=0.7\textwidth]{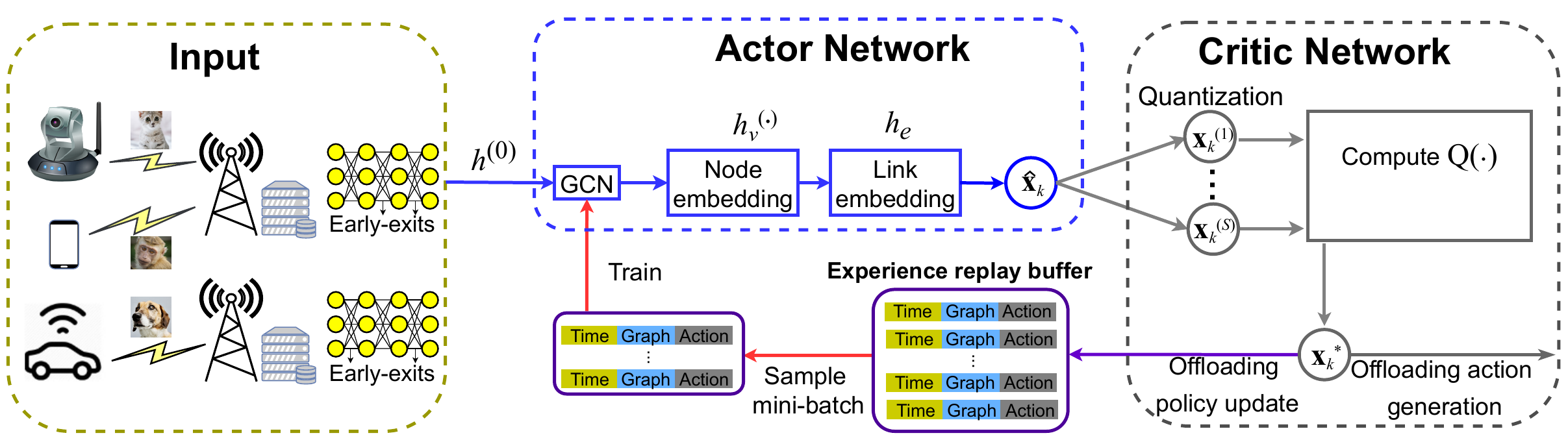}
    \caption{Framework of graph reinforcement learning-based offloading}
    \label{fig:graph}
    \vspace{-2mm}
\end{figure*}
\begin{figure*}[]
\begin{equation}
\small
t_{m,n}^{\textit{w}}\left(k\right)  = \mathop {\arg \max }\limits_{k' \in \mathbb{K},m' \in \mathbb{M}} \left\{ {{\bf{1}}\left(T_{m,n}^{\textit{a}}\left(k\right)-T_{m',n}^{\textit{a}}\left(k'\right) \right) \cdot \left(\underbrace{T_{m',n}^{\textit{a}}\left(k'\right) + t_{m',n}^{\textit{w}}\left(k'\right)+t_{m',n}^{\textit{cmp}}\left(k'\right)}_{\text{Completion time instant of } m'{\text{'s task}}} - T_{m,n}^{\textit{a}}\left(k\right) \right)} \right\}.
\end{equation}
\vspace{-3mm}
\end{figure*}
\section{Problem Formulation}\label{problem}
\subsection{Completion time of an inference task}
There are three steps to complete computation offloading. First, an IoT device sends an inference task to an ES over uplink wireless channel, and then ES performs inference. Finally, ES sends the inference result back to IoT device via downlink wireless channels. Therefore, completion time of an inference task includes communication time, waiting time and computation time, which is denoted as
\begin{equation}
    t_{m,n}\left(k\right) = t_{m,n}^{\textit{com}}\left(k\right) + t_{m,n}^{\textit{w}}\left(k\right) + t_{m,n}^{\textit{cmp}}\left(k\right).
\end{equation}
\subsection{Objective}
In dynamic MEC scenarios, task information of each IoT device and the topology of network may change over time, such as the time-varying task size and wireless channel states, connection establishment and disconnection between IoT devices and ESs. To trade off communication, computation and inference accuracy, we define a reward function, $\textit{Q}\left(\mathcal{G}_k,{\bf{x}}_k\right)$, to denote the achieved reward at time slot $k$, as below
\begin{equation}
    \textit{Q}\left(\mathcal{G}_k,{\bf{x}}_k\right) =\sum\limits_{m \in \mathbb{M}} \sum\limits_{n \in \mathbb{N}} {\Phi_{m,n}\left(k\right) \psi \left(t_{m,n}\left(k\right) \right)} 
    \label{eq:reward}
\end{equation}
where $\mathcal{G}_k$ is the MEC state including task information $\theta_{m,n}^k$ of IoT devices and the computation information $t_{n,l}^{\textit{cmp}}$ and  $\phi_{l}$ of early-exits; ${\bf{x}}_k \triangleq \left\{ \alpha_{m,n}^k \beta_{m,n,l}^k | m \in \mathbb{M},n \in \mathbb{N}, l \in \mathbb{L} \right \}$ is the offloading decision determining to offload device $m$'s task to ES $n$, $\alpha_{m,n}^k$, and ES $n$ performs the task until early-exit $l$, $\beta_{m,n,l}^k$; $\psi \left(x\right)  \triangleq  1-{\textit{Sigmod}}\left(\frac{5x}{\delta_{m,n}^k }\right)$ with
$\psi \left(t_{m,n}\left(k\right)\right) \to 0$ when the completion time $t_{m,n}\left(k\right)$ exceeds the deadline $\delta_{m,n}^k$ and $\psi \left(t_{m,n}\left(k\right)\right) \to 1$ when $t_{m,n}\left(k\right) \to 0$; ${\textit{Sigmod}}\left(\cdot \right)$ is a nonlinear function with the value between 0 and 1 \cite{Giovanni2017}.

The goal of CNN inference offloading is to maximize the average inference accuracy and throughput of inference tasks in a long term perspective, which is formulated as bellows:
\begin{equation}
\max \frac{1}{K} 
\sum\limits_{k \in \mathbb{K}} \textit{Q}\left(\mathcal{G}_k,{\bf{x}}_k\right)
\end{equation}
subject to (\ref{eq2}), (\ref{eq3}) and
\begin{equation}
    t_{m,n}\left(k \right) \leq \delta_{m,n}^k, \forall m \in \mathbb{M} \; {\rm{and}} \; n \in \mathbb{N}
\end{equation}
\section{Graph reinforcement learning-based early-exit mechanism}
\subsection{Candidate early-exits}
At time slot $k$, the computational complexity of offloading decision ${\bf{x}}_k$ is $MN L$, which is huge for MEC scenario when the number of IoT devices and the possible early-exists are large. In fact, it is not necessary to perform the computation with all early-exits. In Section \ref{section:early_exit}, our experiment shows that some exits located deeper in a CNN does not improve inference accuracy. Therefore, we select the five meaningful early-exits in VGG-16 \cite{004} in our experiments instead of using all early-exits (i.e., $L = 17$).
\subsection{GRLE framework}
Fig. \ref{fig:graph} presents the framework of GRLE, where an {\textit{actor-critic}} network is used to generate offloading decisions and update offloading policies. The actor network is responsible for predicting actions; critic network quantifies the prediction and generates offloading decisions; and experience replay buffer stores historical experiences and samples mini-batch training data to train the graph convolutional network (GCN) \cite{Mathias2016}.  

In DROO \cite{Huang2020}, a DNN is used to construct the actor network to extract the features from the input data. However, DNN can only process regular Euclidean data, which is not applicable to dynamic MEC network with graph-like structure data. To better process such data, we apply GCN to replace DNN and analyze the characteristics of graph data through message passing and aggregation between nodes. GCN learns the aggregation method that relies on the relationship between nodes. For a new topology, GRLE does not need to re-train the aggregation function because it can automatically filter the messages from disconnected nodes by graph data updates. As such, GRLE has robust adaptability in dynamic MEC. 

\subsection{Actor network}
Since each task can only be performed by one ES using one early-exit, the two-step task offload decision for device $m$'s task can be merged into one step, i.e., a device has options of $NL$ early-exits to offload its task. We model the structure information of MEC scenario as graph data $\mathcal{G}=\left(\mathcal{V},\mathcal{E}\right)$. In GRLE, $M$ devices and $NL$ early-exits are represented by the graph vertices $\mathcal{V}$, and each IoT device and early-exit is connected by a directed edge $e \in \mathcal{E}$. At time slot $k$, we parse the MEC state $\mathcal{G}_k$ as the initial input data $h^{\left(0\right)}$ for GCN. GCN uses multiple graph convolutional layers to aggregate the neighborhood information. For each node $v \in \mathcal{V}_k$, we define the neighborhood information aggregation
process as follows,
\begin{equation}
    h_v^{\left(i+1\right)}= {\textit{Relu}}\left(W^{\left(i+1\right)} {\textit{C}}\left(h_v^{\left(i\right)},{\textit{A}}^{\left(i\right)}{\left(h_u^{\left(i\right)},u \in N_v \right)}\right)\right)
    \label{eq:aggregation}
\end{equation}
where $W^{\left(i+1\right)}$ is the weight parameters, ${\textit{A}}^{\left(i\right)}\left(\cdot \right)$ is the aggregation function, $N_v$ is the set of node $v$'s neighbors, ${\textit{C}}\left(\cdot \right)$ is a concatenate operation, $\textit{Relu}\left(\cdot \right)$ is a non-linear function \cite{Giovanni2017}. 

The system can acquire the information of tasks and the status of ESs by aggregating the information in the second-order neighborhood of nodes. For example, IoT device $m$ can grasp the information of its second-order neighborhood (other IoT devices connected to ES $n$) through its first-order neighborhood
ES $n$; ES $n$ can acquire the status of its second-order neighborhood (other ESs) through its first-order neighborhood
(IoT devices connected to ES $n$). Therefore, we use two GCN layers in GRLE, i.e., $i \in \left\{0,1\right\}$ in (\ref{eq:aggregation}).

After completing the information aggregation of nodes, we first concatenate the features of the source node (IoT device) and destination node (an early-exit of an ES), $h_e^{\textit{src}}$ and $h_e^{\textit{dst}}$, to get the feature of edge $e \in \mathcal{E}_k$ as
\begin{equation}
    h_e = {\textit{C}}\left(h_e^{\textit{src}}, h_e^{\textit{dst}} \right).
\end{equation}

Then, we can classify the edges to get the relaxed offloading action ${\bf{\hat x}}_k = \left\{{\hat x}_{k,e}|{\hat x}_{k,e}=  {\textit{f}} \left( {h_e} \right), e \in \mathcal{E}_k \right \}$, by the function 
\begin{equation}
    {\textit{f}}\left(h_e\right) = {\textit{Sigmod}}\left({\textit{MLP}}_2\left({\textit{Relu}}\left({\textit{MLP}}_1\left(h_e\right)\right)\right)\right)
\end{equation}
where ${\textit{MLP}}_1$ and ${\textit{MLP}}_2$ are multi-layer perceptions to extract the feature of edge $e$. We use $\textit{Sigmod}\left(\cdot \right)$ function to make the relaxed offloading action satisfy $0 < {\hat x}_{k,e} < 1$ \cite{Giovanni2017}. 
\subsection{Critic network}
We use the order-preserving method in DROO \cite{Huang2020} to quantify the relaxed offloading action ${\bf{\hat x}}_k$ and generate $S$ candidate binary offloading decisions $\mathbb{X}_k = \left\{ {\bf{x}}_k^{\left(1\right)},{\bf{x}}_k^{\left(2\right)},\cdots,{\bf{x}}_k^{\left(S\right)} \right \}$, where ${\bf{x}}_k^{\left(s\right)} = \left\{ x_{k,e}^{\left(s\right)}|x_{k,e}^{\left(s\right)} \in \left\{0,1\right\}, e \in \mathcal{E}_k \right \}$ and $S=MNL$.

Recall that each candidate offloading action ${\bf{x}}_k^{\left(s\right)}$ can achieve reward by solving (\ref{eq:reward}). Therefore, the optimal offloading action at $k\rm{th}$ time slot can be generated as
\begin{equation}
    {\bf{x}}_k^* =\arg \mathop {\max }\limits_  {{\bf{x}}_k^{\left(s\right)} \in \mathbb{X}_k} {\textit Q} \left(\mathcal{G}_k,{\bf{x}}_k^{\left(s\right)}\right).
\end{equation}
\subsection{Offloading policy update}
We use the experience replay buffer technique to train the GCN using the stored data samples $\left(k,\mathcal{G}_k,{\bf{x}}_k^*\right)$, as shown in Fig. \ref{fig:graph}. At time slot $k$, we randomly select a mini-batch of training data $\Delta_k = (\Delta_k^\mathcal{T}, \Delta_k^\mathcal{G}, \Delta_k^\mathcal{X})$ from the memory to update the parameters of GCN and reduce the averaged cross-entropy loss \cite{Huang2020}, as
\begin{equation}
\small
    \xi \left(\Delta_k\right) \hspace{-0.5mm} = \hspace{-1mm} -\frac{1}{|\Delta _k|}\hspace{-1mm}\sum\limits_{k' \in {\Delta_k^\mathcal{T}}} \hspace{-1.5mm}{\bf{x}}_{k'}^*\log {\textit{f}}\hspace{-0.5mm} \left( \mathcal{E}_{k'} \right)
    \hspace{-0.5mm}+\hspace{-0.5mm}\left(1\hspace{-1mm}-\hspace{-1mm}{\bf{x}}_{k'}^*\right) \log \left(1\hspace{-1mm}-\hspace{-0.5mm}{\textit{f}}\hspace{-0.5mm} \left( \mathcal{E}_{k'} \right)\right) 
\end{equation}
where $|\Delta _k|$ is the size of the mini-batch training data, $\Delta_k^\mathcal{T}$ is the set of time slots, $\Delta_k^\mathcal{G}$ is the set of graphs, and $\Delta_k^\mathcal{X}$ is the set of actions. The detailed process of GRLE is described in Algorithm 1.
\begin{algorithm}
\caption{GRLE for offloading decision-making}
\KwIn{MEC state $\mathcal{G}_k, \forall k \in \mathbb{K}$, training interval $\omega$}
\KwOut{Offloading decision ${\bf{x}}_k^*$}
\For{$k =1,2,\cdots,K$}{
Generate the relaxed offloading action ${\bf{\hat x}}_k$ in (14);\\
Quantify ${\bf{\hat x}}_k$ into $S$ binary actions $\mathbb{X}_k$ ;\\
Select the optimal offloading action ${\bf{x}}_k^*$ using (15);\\
Update the experience
replay buffer by adding $\left(\mathcal{G}_k,{\bf{x}}_k^*\right)$.\\
\If{$k$ \rm{mod} $\omega$ = 0}{
Randomly sample a mini-batch of training data $\Delta _k$ from the buffer,
train GCN and update the parameters using (16).
}}
\label{alg1}
\end{algorithm}
\section{Performance Evaluation}\label{simulation}
\subsection{Simulation setup and methodology}
In this experiment, we consider an MEC network with 14 IoT devices and 2 ESs (RTX 2080TI and GTX 1080TI). We consider image recognition task using VGG-16, which should be processed within a fixed deadline of $\delta_{m,n}^k=30$ms. For IoT device $m$, the task size $d_{m,n}^k$ is between 50 KBytes and 100 KBytes, and the uplink transmission data rate $r_{m,n}^k$ is between 20 Mbps and 100 Mbps. We implement GRLE with PyTorch, and the training parameters are set as: the hidden neurons of two GCN layers are 128 and 64; the learning rate is initialized as 0.001; the experience replay buffer size is 128; the mini-batch size $|\Delta _k| = 64$; the training interval $\omega=10$; the optimizer of loss function $\xi \left(\Delta_k\right)$ is Adam function \cite{004}. To validate the effectiveness of our proposed GRLE, we compare it with the following  three offloading methods:
\begin{itemize}
    \item GRL: Basic graph reinforcement learning; 
    \item DROO: The state-of-the-art work \cite{Huang2020};
    \item DROOE: DROO with early-exit mechanism.
\end{itemize}
\subsection{Performance of early-exits} \label{section:early_exit}
\begin{figure*}
\centering
\begin{minipage}[]{0.34\textwidth}
\centering
\captionsetup{type=table}
\vspace{-12mm}
\caption{\\Candidate early-exits}
\centering
\scalebox{0.7}{
\begin{tabular}{c|c|cc}
\hline
\multirow{2}{*}{No. of early-exit} & \multirow{2}{*}{Acc.} & \multicolumn{2}{c}{Inference   time (ms)}                          \\ \cline{3-4} 
                                     &                       & \multicolumn{1}{l|}{RTX\_2080TI} & \multicolumn{1}{l}{GTX\_1080TI} \\ \hline
1                                    & 0.8                   & \multicolumn{1}{c|}{0.36}        & 0.73                             \\ 
3                                   & 0.85                  & \multicolumn{1}{c|}{0.46}        & 0.89                             \\ 
4                                    & 0.885                 & \multicolumn{1}{c|}{0.54}        & 1.06                             \\ 
7                                    & 0.905                 & \multicolumn{1}{c|}{0.71}        & 1.4                              \\ 
17                                    & 0.935                 & \multicolumn{1}{c|}{1.26}        & 2.42                             \\ \hline
\end{tabular}}
\label{lab:candidates}
\end{minipage}
\hspace{1mm}
\begin{minipage}[]{0.3\textwidth}
    \centering
    \includegraphics[width=1\textwidth]{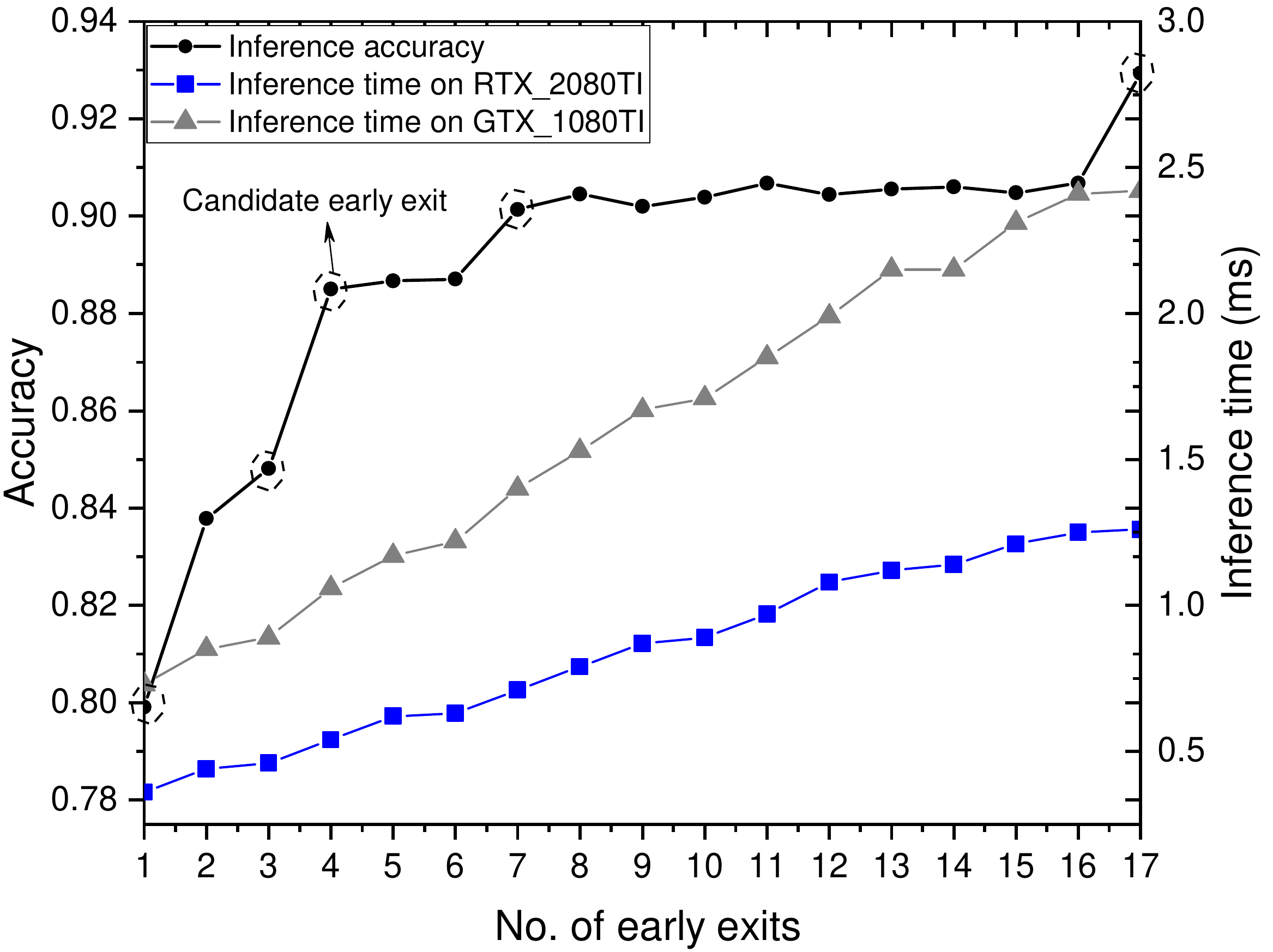}
    \vspace{-6mm}
    \caption{Performance of early-exits}
    \label{fig:acc}
\end{minipage}
\hspace{0.1mm}
\begin{minipage}[]{0.3\textwidth}
\centering
    \includegraphics[width=1\textwidth]{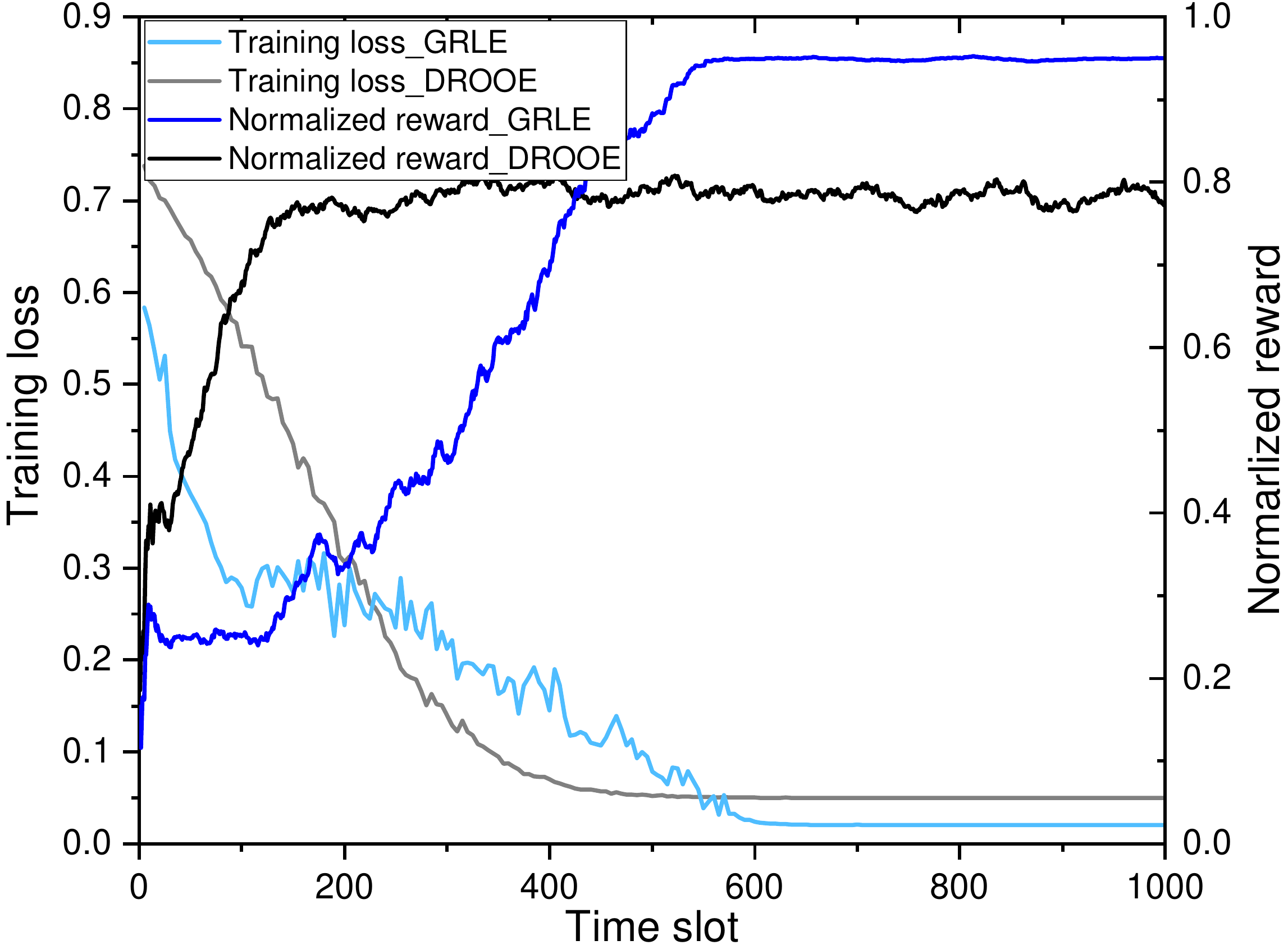}
    \vspace{-5mm}
   \caption{Performance of convergence}
    \label{fig:convergence}
\end{minipage}
\vspace{-3mm}
\end{figure*}

In this experiment, we first train the main branch of VGG-16 on the dataset cifar-10 \cite{Alex2009}, and then use it as the pre-trained model to train the sub-models of early-exits. We add a classifier after each CL or pooling layer as an early-exit, where early-exit 17 is the main branch. As shown in Fig. \ref{fig:acc}, the exits located deeper in the CNN model do not always increase inference accuracy. This means that we need to select the meaningful exits, which is also necessary to reduce the computation complexity for offloading decision. Here we choose the five circled points in Fig. \ref{fig:acc} as the candidate early-exits of VGG-16. The detailed information of the candidate early-exits is summarized in Table \ref{lab:candidates}.
\begin{figure*}
    \centering
    \subfigure[Average accuracy]{
    \includegraphics[width=0.29\textwidth]{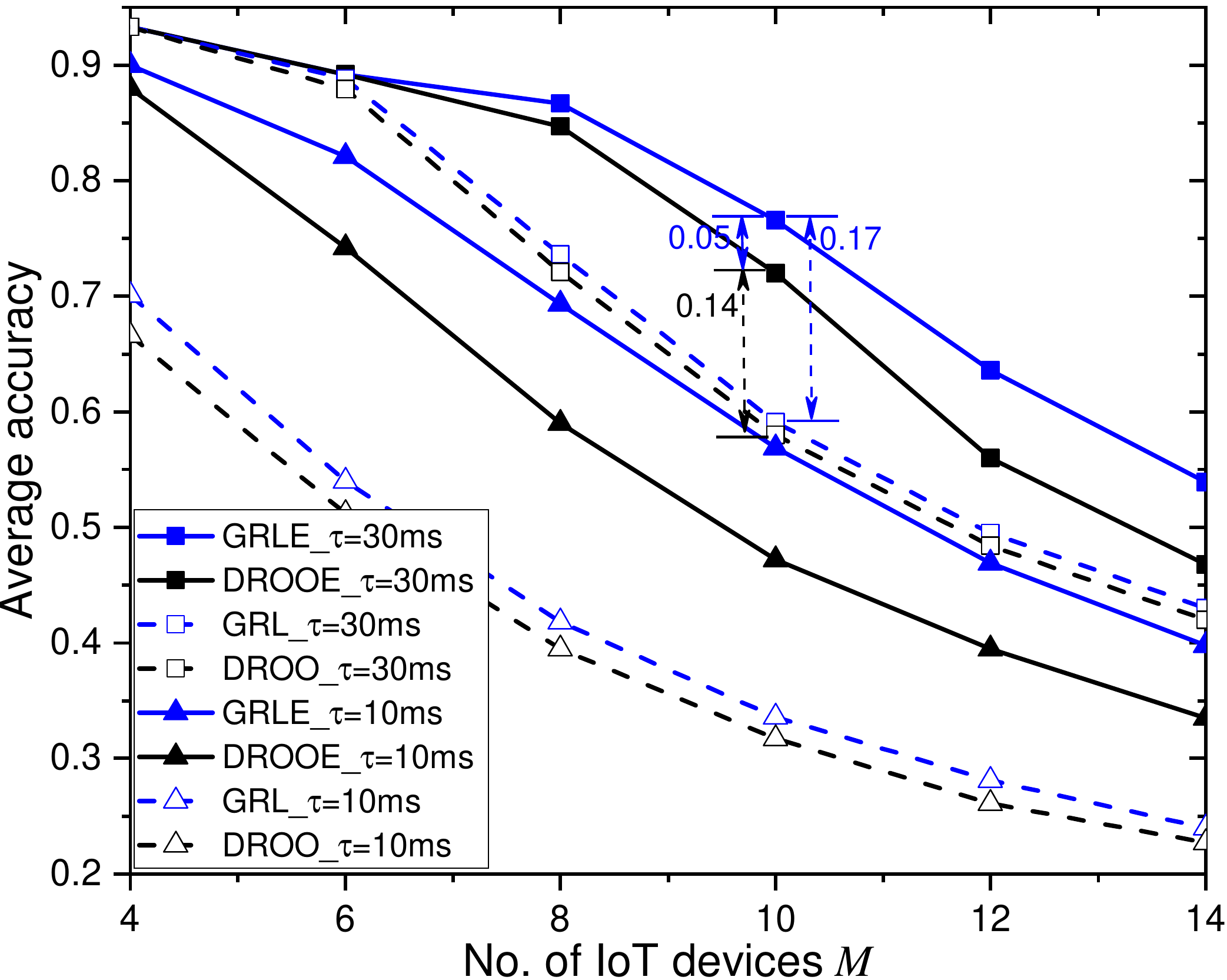}
    \label{full_acc}}
    \hspace{2mm}
    \subfigure[Service successful probability]{
    \includegraphics[width=0.3\textwidth]{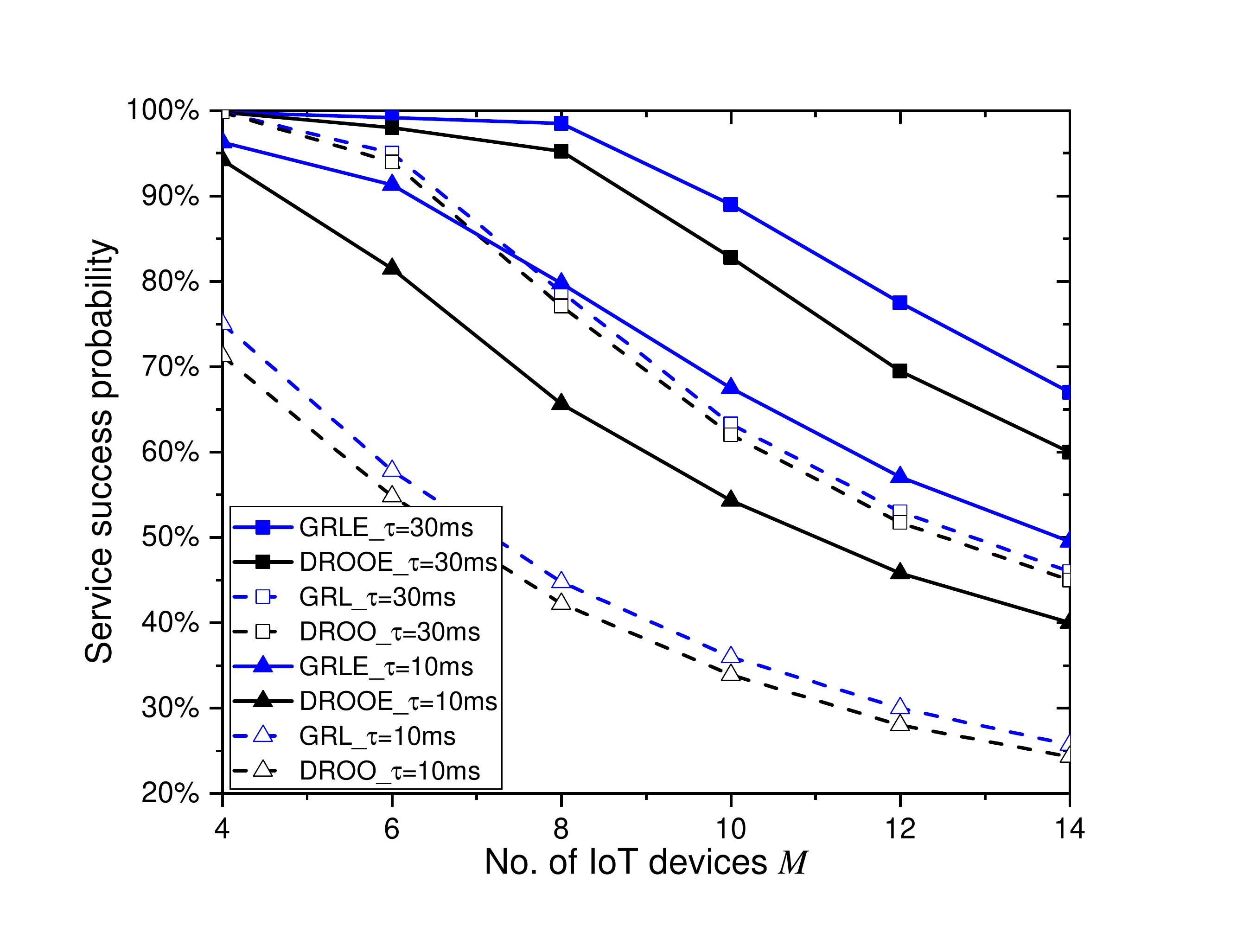}}
    \hspace{2mm}
    \subfigure[Average throughput]{
    \includegraphics[width=0.29\textwidth]{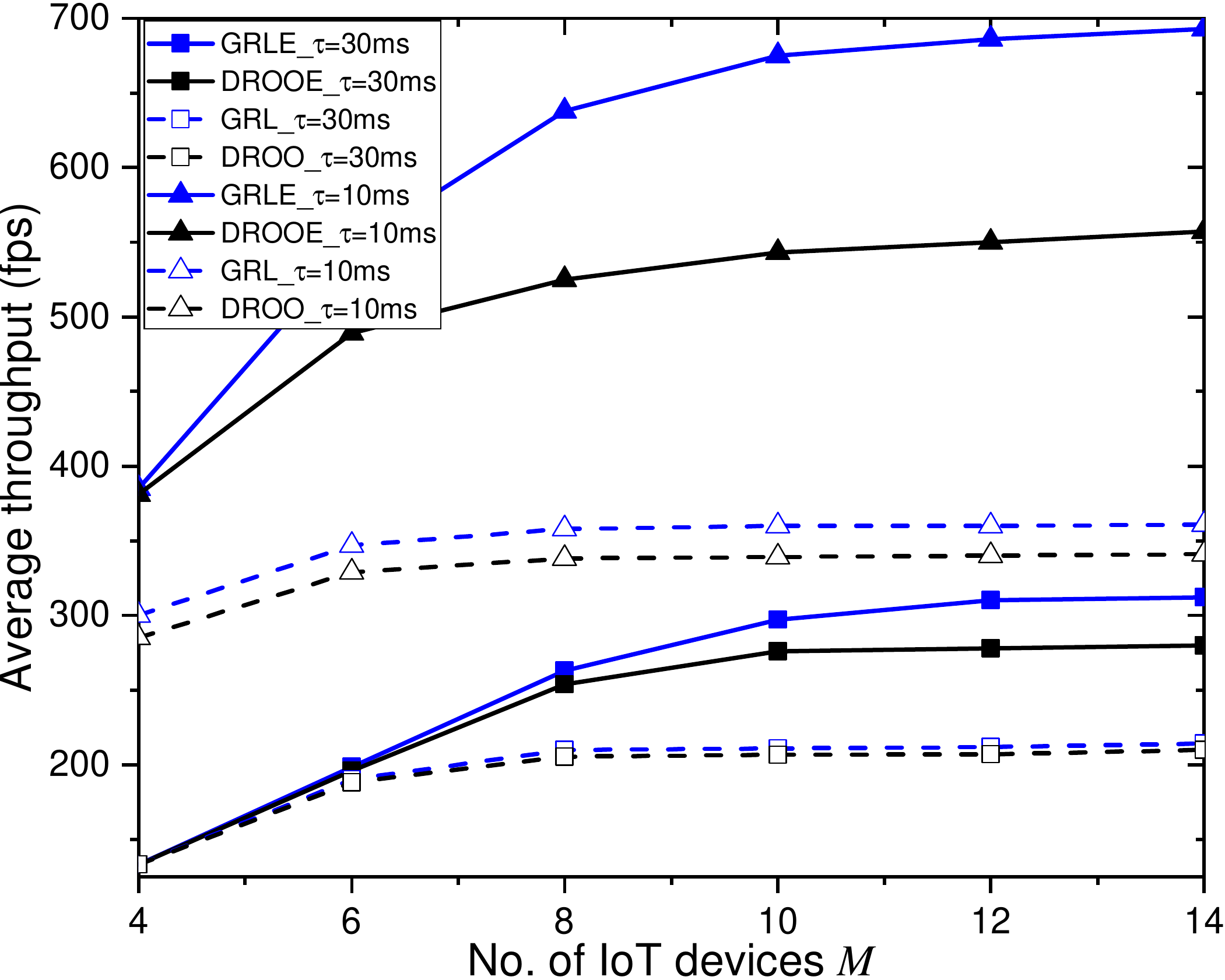}}
    \vspace{-3mm}
    \caption{Performance under various No. of IoT devices}
    \label{fig:iot}
\vspace{-4mm}
\end{figure*}
\subsection{Performance of Convergence}
We first define the normalized reward for time slot $k$ as
\begin{equation}
    {\hat {\textit{Q}}} \left(\mathcal{G}_k,{\bf{x}}_k\right) = \frac{{\textit{Q}} \left(\mathcal{G}_k,{\bf{x}}_k \right)  }{{\textit{Q}} \left(\mathcal{G}_k,{\bf{x'}}_k\right) }
\end{equation}
where the action ${\bf{x'}}_k$ is obtained by exhaustive searching.

In Fig. \ref{fig:convergence}, we plot the moving average of ${\hat {\textit{Q}}}$ over the last 50 time slots and training loss. As the time slot increases, the moving average of ${\hat {\textit{Q}}}$ and training loss gradually converge to the optimal solution, where the fluctuations are mainly due to the random sampling of training data. Specifically, the moving average of the normalized reward ${\hat {\textit{Q}}}$ of GRLE exceeds 0.96 and the training loss is below 0.03, which outperforms that of DROOE. This is because GRLE makes full use of MEC states for offloading decisions, while DROOE only considers the wireless channel states.
\subsection{Performance evaluation and analysis}
We test the performance of different offloading methods for 10,000 time slots in different scenarios. We define a successful task as the one that is completed within the deadline. The service success probability (SSP) is the ratio of the number of all successful tasks to the total number of tasks. The average inference accuracy is the sum of the achieved accuracy of successful tasks divided by the total number of tasks. The average throughput is the number of successful tasks divided by the sum of all time slots.
\subsubsection{Performance under changing number of IoT devices}
As shown in Fig. \ref{fig:iot}, the average accuracy and SSP decreases as $M$ increases for the given ES computation resources. This is because when $M$ is large, more tasks fail to meet their deadlines due to the limited resources of ESs. As such, the average throughput gradually reaches a plateau. Moreover, the system achieves higher throughput at $\tau=10$ms than that of $\tau=30$ms; however, its average accuracy and SSP are lower. This is because the higher task generation rate at $\tau=10$ms allows the system to make better use of ES's idle time to process more tasks in the same time; however, this will result in more failed tasks because of the higher occupancy of ESs and wireless channels. In addition, using early-exits it can improve average accuracy, SSP and throughput, especially when $M$ is large. This is because a task that fails to meet the deadline using static inference can be terminated earlier to meet
their application deadlines using early-exits. For example, with early-exits the average accuracy of GRLE is improved by 0.17 compared with GRL, when $M=10$ and $\tau=30$ms. Compared with DROOE, GRLE achieves better performance because GRLE can use the full information of MEC to make offloading decisions.
\subsubsection{Performance under various available capacity of ESs}
In real scenarios, ES is not always completely idle at each time slot. Therefore, we assume that each ES has a stochastic available computation resource between 25\% and 100\% of the total computational capacity at each time slot. Compared with the average accuracy in Fig. \ref{full_acc}, Fig. \ref{fig:ES} shows that GRLE and DROOE achieve more improvement by using early-exits. For example, when $M=10$ and $\tau=30$ms, GRLE improves the average accuracy up to 0.56, equivalent to 2.8$\times$ (i.e., $0.56/0.2$) over GRL and DROOE improves the average accuracy up to 0.41, equivalent to 2.28$\times$ (i.e., $0.41/0.18$) over DROO in Fig. \ref{fig:ES}, which is larger than that of 1.28$\times$ (i.e., $0.77/0.6$) and 1.24$\times$ (i.e., $0.72/0.58$) in Fig. \ref{full_acc}. Meanwhile, GRLE achieves the average accuracy of 1.37$\times$ (i.e., $0.56/0.41$) over DROOE, which outperforms that of 1.07$\times$ (i.e., $0.77/0.72$) in Fig. \ref{full_acc}. When $\tau=10$ms, the gain of average accuracy thanks to the early-exits in Fig. \ref{fig:ES} are larger than that in Fig. \ref{full_acc}. This is because early-exit mechanism allows ESs with limited computation resource to terminate the computation earlier to reduce the number of failed tasks and GRLE can learn the state information of ESs better to make offloading decision.
\subsubsection{Performance under various fluctuation of inference time}
\begin{figure*}
\centering
\begin{minipage}[]{0.29\textwidth}
    \centering
    \includegraphics[width=1\textwidth]{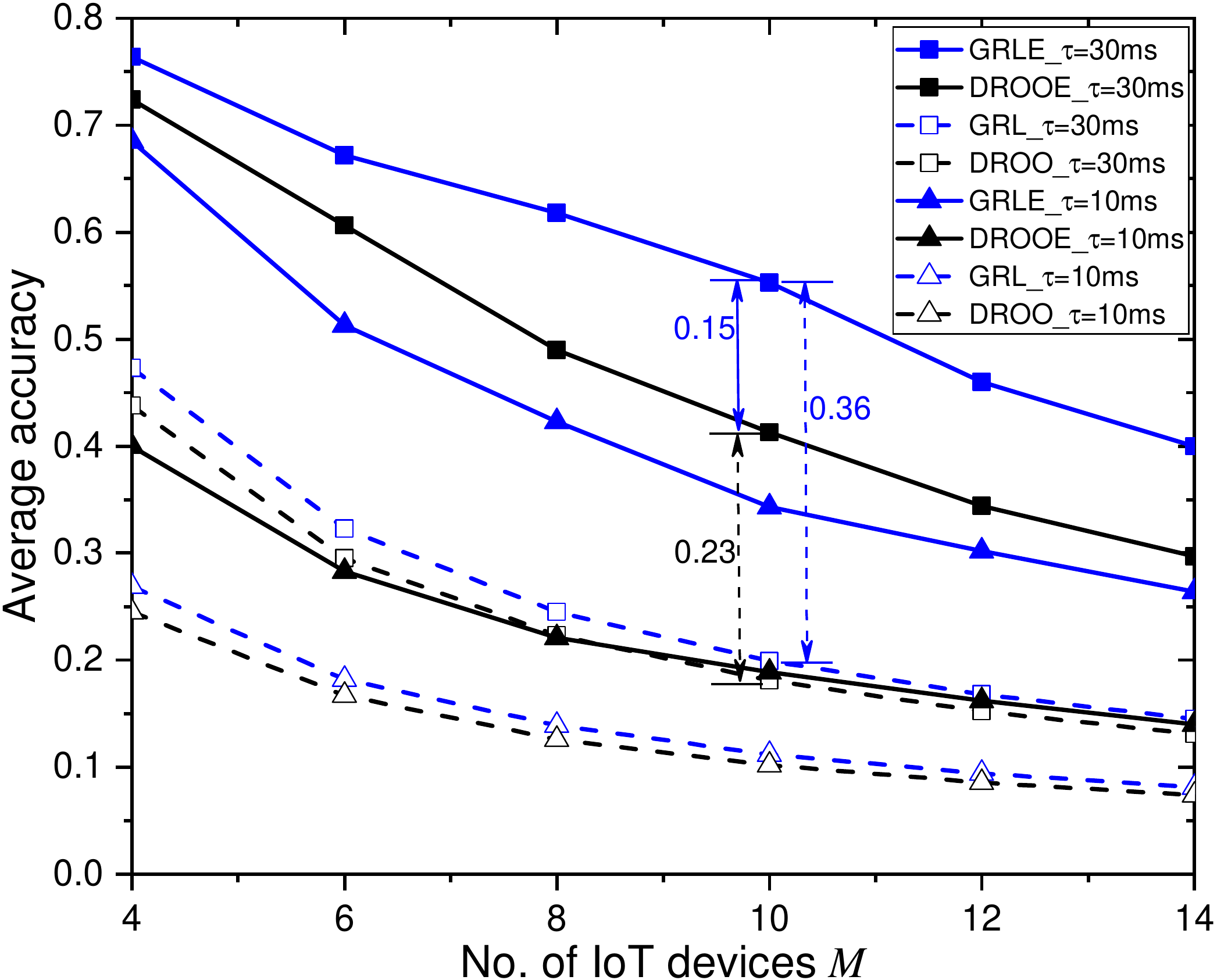}
    \vspace{-5mm}
    \caption{Performance under various available capacity of ESs}
    \label{fig:ES}
\end{minipage}
\hspace{3mm}
\begin{minipage}[]{0.29\textwidth}
\centering
    \includegraphics[width=1\textwidth]{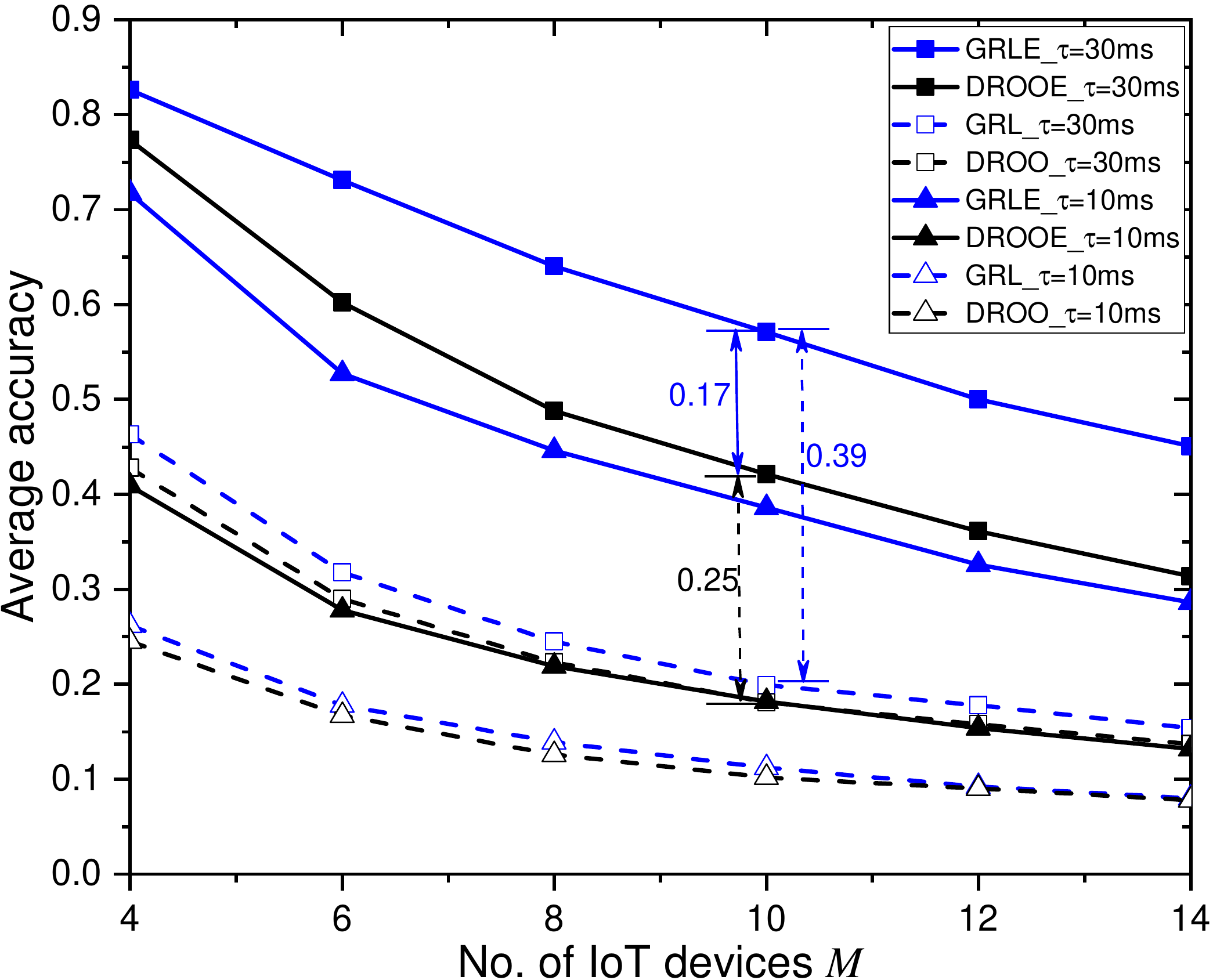}
    \vspace{-5mm}
    \caption{Performance under various fluctuation of inference time}
    \label{fig:fluctuation}
\end{minipage}
\hspace{3mm}
\begin{minipage}[]{0.29\textwidth}
    \centering
    \includegraphics[width=0.99\textwidth]{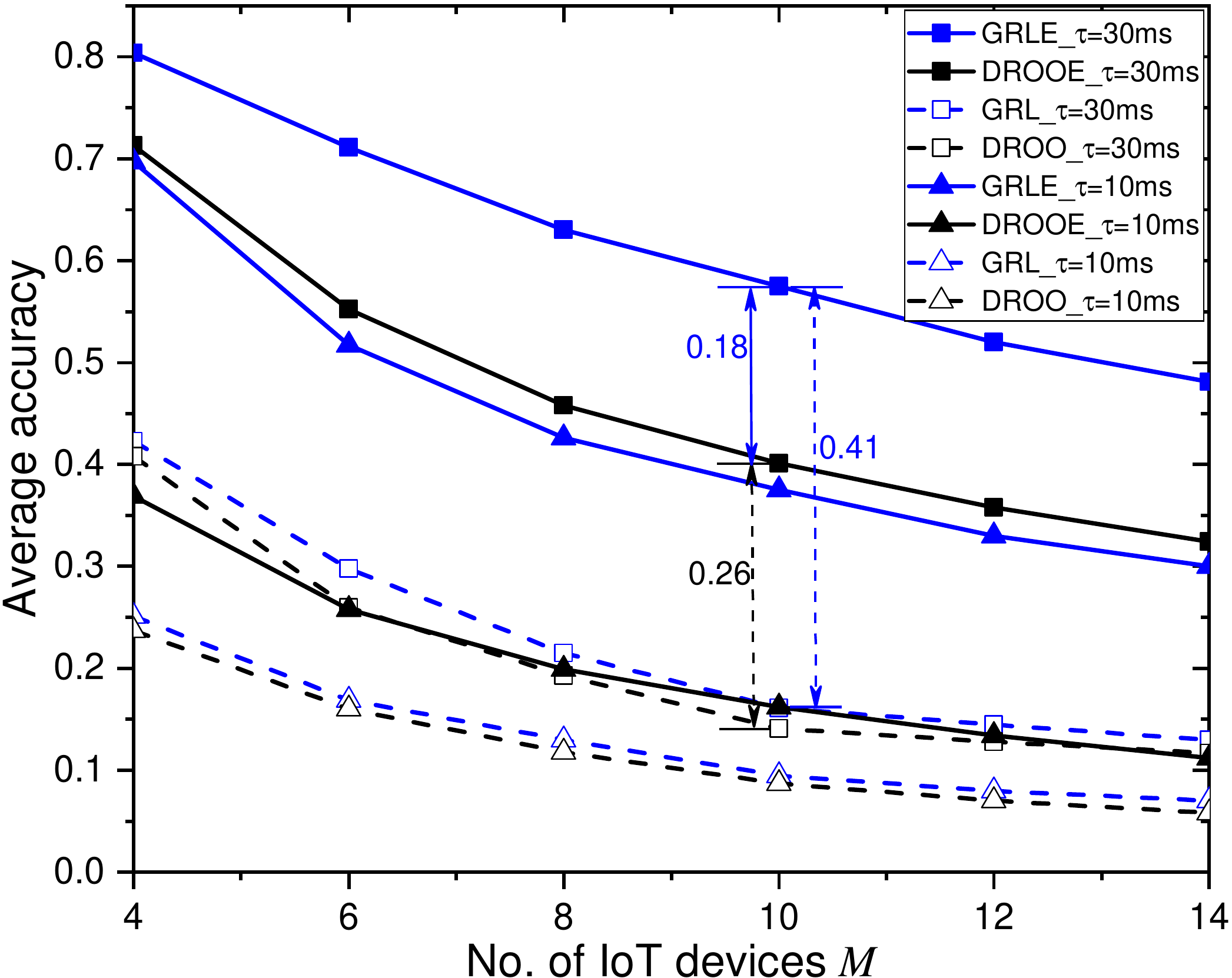}
    \vspace{-5mm}
    \caption{Performance under imperfect channel state information}
    \label{fig:csi}
\end{minipage}
\vspace{-4mm}
\end{figure*}

In addition to the variation of ES computation resource, we considered a realistic case where the inference time of each early-exit fluctuates $\pm 25\%$ of its measured value in Table \ref{lab:candidates}. In Fig. \ref{fig:fluctuation}, GRLE achieves even higher improvement in average accuracy over other offloading methods. For example, GRLE achieves the average accuracy of 2.86$\times$ (i.e., $0.6/0.21$) over GRL and 1.39$\times$ (i.e., $0.6/0.43$) over DROOE when $M=10$ and $\tau=30$ms. This is because GRLE can better learn the dynamic environment of MEC, and the early-exit mechanism ensures that more tasks are processed within the deadline.
\subsubsection{Performance under imperfect channel state information}
Since channel estimation is often not perfect in practical systems, in this study we include the transmission fluctuation of $ \pm 20\%$  of the estimated values, in addition to the variation of ES computation resource and the variation of inference time. As shown in Fig. \ref{fig:csi}, GRLE achieves a higher improvement in average accuracy compared to other offloading methods. For example, in Fig. \ref{fig:csi}, GRLE achieves the average accuracy of 3.41$\times$ (i.e., $0.58/0.17$) over GRL and 1.45$\times$ (i.e., $0.58/0.4$) over DROOE when $M=10$ and $\tau=30$ms.
This is because GRLE can better aggregate all the information of dynamic MEC, including the uncertainty in communication time and the stochastic available capacity of ESs, and then select better early-exits to perform computation to ensure that more tasks are processed within the deadline. This further demonstrates the robust adaptability of GRLE for offloading decision-making 
in dynamic MEC scenarios.
\section{Conclusion}\label{conclusion}
This paper studies computation offloading of CNN inference tasks in MEC networks. We use early-exit mechanism to address the uncertainties
in communication time and available computation resource at ESs by terminating the inference earlier to meet the deadline with a slight compromise of inference accuracy. We design a reward function to trade off the inference accuracy and completion time, and formulate the CNN inference offloading problem as a maximization problem to maximize the average inference accuracy and throughput in long term. To solve this problem, we propose GRLE to make effective offloading decisions in dynamic MEC.

\ifCLASSOPTIONcaptionsoff
  \newpage
\fi

\section*{Acknowledgment}
This work is supported by Agile-IoT project (Grant No. 9131-00119B) granted by the Danish Council for Independent Research.

\bibliographystyle{IEEEtran}
\bibliography{ddnn}

\end{document}